\documentclass[journal]{IEEEtran}
\usepackage{array}
\usepackage[caption=false,font=normalsize,labelfont=sf,textfont=sf]{subfig}
\usepackage{textcomp}
\usepackage{verbatim}

\def\BibTeX{{\rm B\kern-.05em{\sc i\kern-.025em b}\kern-.08em
    T\kern-.1667em\lower.7ex\hbox{E}\kern-.125emX}}
\usepackage{balance}

\usepackage{amssymb}
\usepackage{amsthm}
\usepackage{amsmath}
\usepackage{amsfonts}
\usepackage{multirow}
\usepackage{algorithm}
\usepackage{algorithmic}
\usepackage{tcolorbox}
\usepackage{tabularx,colortbl}
\usepackage{adjustbox}
\usepackage{hhline}% http://ctan.org/pkg/hhline
\usepackage{amssymb} % assumes amsmath package installed
\usepackage{mathtools} 
\usepackage{bm}
\usepackage{enumitem}
\usepackage{multirow}
\usepackage{epsfig}
\usepackage{epstopdf}
\usepackage{graphicx}
\usepackage{graphics}
\usepackage{hyphenat}
\usepackage{fancyhdr}
\usepackage{lipsum}
\usepackage{stfloats}
\usepackage{indentfirst}
\usepackage[english]{babel} % English language/hyphenation
\usepackage{cleveref}
\usepackage{xcolor}
\usepackage{url}

\usepackage{graphicx}

\makeatletter
\DeclareFontFamily{U}{tipa}{}
\DeclareFontShape{U}{tipa}{m}{n}{<->tipa10}{}
\newcommand{\arc@char}{{\usefont{U}{tipa}{m}{n}\symbol{62}}}%

\newcommand{\arc}[1]{\mathpalette\arc@arc{#1}}

\newcommand{\arc@arc}[2]{%
  \sbox0{$\m@th#1#2$}%
  \vbox{
    \hbox{\resizebox{\wd0}{\height}{\arc@char}}
    \nointerlineskip
    \box0
  }%
}
\usepackage{mathtools}
\DeclarePairedDelimiter\ceil{\lceil}{\rceil}
\DeclarePairedDelimiter\floor{\lfloor}{\rfloor}
\makeatother

\begin{document}

\title{A Decentralized Spike-based Learning Framework for Sequential Capture in Discrete Perimeter Defense Problem}
\author{~Mohammed~Thousif\IEEEauthorrefmark{1},~Shridhar~Velhal\IEEEauthorrefmark{1},~Suresh~Sundaram,~and~Shirin~Dora
\thanks{Thousif, Shridhar Velhal, and Suresh Sundaram are with the Department of Aerospace Engineering, Indian Institute of Science, Bengaluru, India. Email: pagalat@iisc.ac.in, velhalb@iisc.ac.in, vssuresh@iisc.ac.in}
\thanks{Shirin Dora is with the Department of Computer Science, Loughborough Univerity, United Kingdom. Email: s.dora@lboro.ac.uk}}

\maketitle
\markboth{}{}
\begingroup\renewcommand\thefootnote{\IEEEauthorrefmark{1}}
\footnotetext{Equal contribution}

\begin{abstract}
This paper proposes a novel Decentralized Spike-based Learning (DSL) framework for the Perimeter Defense Problem (PDP). The PDP in this paper is termed discrete-PDP (d-PDP), as the circular territory is discretized into multiple segments. A team of defenders is operating on the perimeter to protect the circular territory from radially incoming intruders. At first, the d-PDP is formulated as a spatio-temporal multi-task assignment problem (STMTA). The problem of STMTA is then converted into a multi-label learning problem to obtain labels of segments that defenders have to visit in order to protect the perimeter. The DSL framework uses a Multi-Label Classifier using Synaptic Efficacy Function spiking neuRON (MLC-SEFRON) network for deterministic multi-label learning. Each defender contains a single MLC-SEFRON network. Each MLC-SEFRON network is trained independently using input from its own perspective for decentralized operations. The input spikes to the MLC-SEFRON network can be directly obtained from the spatio-temporal information of defenders and intruders without any extra pre-processing step. The output of MLC-SEFRON contains the labels of segments that a defender has to visit in order to protect the perimeter. Based on the multi-label output from the MLC-SEFRON a trajectory is generated for a defender using a Consensus-Based Bundle Algorithm (CBBA) in order to capture the intruders. The target multi-label output for training MLC-SEFRON is obtained from an expert policy. Also, the MLC-SEFRON trained for a defender can be directly used for obtaining labels of segments assigned to another defender without any retraining. The performance of MLC-SEFRON has been evaluated for full observation and partial observation scenarios of the defender. The overall performance of the DSL framework is then compared with expert policy along with other existing learning algorithms. The scalability of the DSL has been evaluated using an increasing number of defenders.  
\end{abstract}


\section{Introduction}

\IEEEPARstart{T}{echnological} advances in sensors and computer vision have enabled organizations to use autonomous Unmanned Aerial Vehicles (UAVs) for various applications such as logistics \cite{kuru2019analysis}, search and rescue \cite{silvagni2017searchrescue}, agriculture \cite{gonzalez2013agriculture}, security and surveillance \cite{chen2007surveillance}, fire-fighting and defence \cite{shakhatreh2019unmanned}. The use of UAVs also pose a threat to the privacy and security of vital infrastructures  \cite{gusterson2016drone}. UAV-based methods for this purpose have been evaluated for the protection of critical  airspace infrastructures in \cite{kang2020protect}. 
These methods employ a team of UAVs, termed as defenders, that patrol the boundary of critical infrastructures to prevent intruders from infiltrating them. This problem is termed as the Perimeter Defense Problem (PDP) \cite{shishika2018local,shishika2020cooperative}. Generally, the defenders are constrained to operate on the perimeter of the territory. The PDP has been addressed for different boundary shapes like  linear territory \cite{smith2009dynamic}, conical territory \cite{bajaj2021competitive}, circular territory \cite{bajaj2019dynamic}, and more generally for any convex territory \cite{shishika2018local,shishika2020cooperative}. A detailed review of existing approaches and challenges for the PDP are presented in \cite{shishika2020review,yan2023multiplayer}.

The differential geometric approaches have been used to  compute feasible regions for defenders. Then using these feasibility constraints, one-to-one assignments of the defenders to intruders are computed as in \cite{shishika2018local,shishika2020cooperative}. However, these geometric approaches are limited to one-to-one capture and require full observation.  In \cite{paulos2019decentralization}, the decentralized policy for communication and decision-making for defenders is obtained from a centralized solution. But this decentralized solution allows each defender to capture only one intruder. As a result, the solution obtained does not account for sequential capture, where a defender captures multiple intruders in a sequence. 
In \cite{macharet2020adaptive}, a two-stage adaptive partitioning approach is proposed for protecting circular territories against multiple radially incoming intruders. The territory is divided into partitions in the first stage of the partitioning approach. In the second stage, intruders in each partition are independently assigned to the defenders. However, the defenders lack cooperation as the defender in each partition solves the assignments independently.

 In \cite{velhal2022decentralized} and \cite{velhal2022DREAM}, PDP has been formulated as a spatio-temporal Multi-Task Assignment Problem (STMTA) for convex territories. In this formulation, each intruder's arrival time and location on the perimeter are predicted by defenders. Then to neutralize an intruder, a defender has to reach that arrival location at the arrival time of that intruder, which represents a spatio-temporal task. The PDP is converted into an STMTA problem when there are multiple spatio-temporal tasks for a single defender to handle. The solution to the STMTA problem is necessary to compute assigned intruders for each defender. The trajectory, following which assigned intruders will be neutralized, is computed for each defender using the assignment solution. In \cite{adler2022role}, the dynamic programming approach has been presented to solve the STMTA formulation of PDP. However, due to its exponential computational complexity, it is not implementable in real-time.

Analytical solutions for STMTA based on the exact time-constrained multiple Traveling Salesman Problem (mTSP) problem have been developed in \cite{chopra2015spatio, velhal2022non} for music-playing robots, in \cite{velhal2023dynamic} for warehouse automation, and in \cite{velhal2022decentralized,velhal2022DREAM} for PDP. These approaches involve solving the linear sum assignment problem using the Hungarian solution for mTSP, which has a computational complexity of order ${(N+M-1)}^3$.  The aforementioned works based on analytical, geometric, and numerical approaches are not scalable. Their complexity grows with an increase in the number of intruders. Hence they are limited to the protection of small territories. Designing decentralized and scalable strategies with partial observability is one of the major challenges for real-time implementations of perimeter defense systems.

Recently, Graph Neural Networks (GNN) are used to develop a learning-based decentralized approach for solving PDP as an assignment problem \cite{lee2023graph}. Each defender is equipped with a vision camera that uses ANN to extract features of observed intruders and defenders. These features are used as input to GNN. The GNN communicates with its neighboring defender (i.e. another GNN) to share features dynamically. The maximum matching algorithm \cite{chen2014multiplayer} is used as an expert policy for training the GNN. The GNN has a fixed number of output neurons representing the closest intruders to a defender (arranged in a specific order).  The choice of the number of input and output channels affects performance. Also,  as the number of input and output channels is hard-coded, the setting is not generic.
However, the GNN approach is only suitable for one-to-one assignments of defenders to intruders.  Thus, there is a need to develop a generic learning-based scalable solution  for one-to-many assignments of defenders to intruders to protect large territories.   

In this paper, a discrete PDP (d-PDP) is considered in which a defender needs to capture an intruder by visiting the segment of that intruder at the arrival of that intruder. 
To the best of the author's knowledge for the first time in this paper, the STMTA problem posed by d-PDP is converted into a deterministic multi-label learning problem using a Spiking Neural Network (SNN). The conventional multi-label learning algorithms provide a probabilistic solution, but multi-task assignment problem posed by d-PDP requires a deterministic solution for better performance. For predicting the multi-label deterministically a novel multi-label classifier using SNN is developed in this paper.
The input to the SNN is obtained from the spatio-temporal information about the defenders and intruders. Each defender is trained using a different SNN, enabling the decentralized operation of the DSL framework. The DSL framework is executed for different sensing ranges of defenders. If a defender is capable of sensing the entire perimeter then it is termed a full observation scenario. Otherwise, it is termed a partial observation scenario.
 
DSL framework is designed to handle the spatio-temporal problem posed by d-PDP. As the d-PDP has inherent spatio-temporal nature, the DSL framework does not require any pre-processing to generate input spikes. The DSL framework is event-triggered which means that a neuron in the framework emits a spike when an object (either a defender or intruder) is detected otherwise, the neuron remains silent. This spiking nature of DSL makes it energy efficient.

For the deterministic multi-label learning, each label prediction is formulated as a binary classification problem using Synaptic Efficacy Function spiking neuRON (SEFRON) \cite{mc-sefron} network. Hence the classifier is termed a Multi-Label Classifier using SEFRON (MLC-SEFRON). A 3-layered MLC-SEFRON architecture is used for this purpose namely input layer, SEFRON layer and output layer. Each label prediction is evaluated using the response of two neurons in the SEFRON layer. If one neuron spikes earlier than the other then the output neuron connected to these two neurons in the SEFRON layer generates a label $1$ and vice versa. If the response of the output neuron is $1$ then the defender is assigned to the segment associated with that neuron and vice versa. Therefore MLC-SEFRON learning predicts the labels in a deterministic manner. Since the number of intruders exceeds the number of defenders considered in this paper, the MLC-SEFRON is designed to predict multiple labels of segments assigned for a single defender.
Each defender has one MLC-SEFRON architecture. The DSL framework is designed such that all the defender uses the same trained MLC-SEFRON architecture.  After training the MLC-SEFRON network with a single defender, it can be used for obtaining assignments of other defenders without retraining. Hence the DSL framework is scalable and can be used for any number of defenders without any extra learning computations.
Once defenders obtain their assigned labels, they may have conflicts in their trajectories due to the decentralized approach. These conflicts are resolved to generate the trajectories of each defender using the Consensus-Based Bundle Algorithm (CBBA).
The dataset and expert assignment labels for training are generated by solving d-PDP using expert policy derived from the simplified form of \cite{velhal2022DREAM}.

The performance of MLC-SEFRON in training is evaluated using two different observation scenarios. One is the partial observation scenario and the other is the full observation scenario of the defender around the perimeter. In a partial observation scenario, the defender is assumed to have a sensing range of 150 degrees of the perimeter. Whereas in full observation, the defender has a sensing range of 360 degrees which covers the entire perimeter. The MLC-SEFRON trained for one defender can be used for other defenders without retraining, hence the learning in this paper is decentralized. The performance of the MLC-SEFRON is evaluated using mutli-label metrics such as $Precision$, $Recall$,  and $F1-Score$ as in \cite{mll-metrics1}.
The DSL framework is trained with five defenders using the MLC-SEFRON algorithm, and then the same is tested. 
The results indicate that the DSL framework's success rate is on par with the expert solution. 
The performance of DSL framework is then compared with the state-of-the-art adaptive partitioning approach \cite{macharet2020adaptive} and the expert policy. The results indicate that the DSL framework performs better compared.
Furthermore, the DSL performance is evaluated for different-sized teams of defenders without retraining, to illustrate the scalability of the DSL framework.  The DSL framework performs at par compared to the centralized expert policy for the different sizes of defenders to showcase its  generalization performance.

 The main contributions of this paper can be summarized as follows:

\begin{enumerate}
    \item The formulation of the d-PDP into the spiking multi-label learning problem with the help of MLC-SEFRON architecture.     
    \item  Development of deterministic MLC-SEFRON learning algorithm for predicting multiple labels of segments.   
    \item Distributed and scalable learning-based solution for sequential capture in d-PDP. To the authors' best knowledge this is the first time in literature, a learning-based solution is proposed for fewer defenders protecting territory against multiple intruders by sequential capture.
\end{enumerate}

The rest of the paper is organized as follows: Section \ref{sec:problem_formulation} presents the mathematical formulation of d-PDP for circular perimeter as a multi-label classification problem using SNN. 
Section \ref{sec:architecture} presents the SNN architecture and MLC-SEFRON learning algorithm.
 Section \ref{sec:result} provides the performance evaluation results for MLC-SEFRON learning algorithms using multi-label metrics. This section also presents results on performance comparison of the DSL framework with expert policy and ablation studies exhibiting scalability. Finally, Section \ref{sec:conclusion}  summarises the conclusions from this paper.

\begin{figure*}[t] 
\centering
\includegraphics[width=0.95\linewidth]{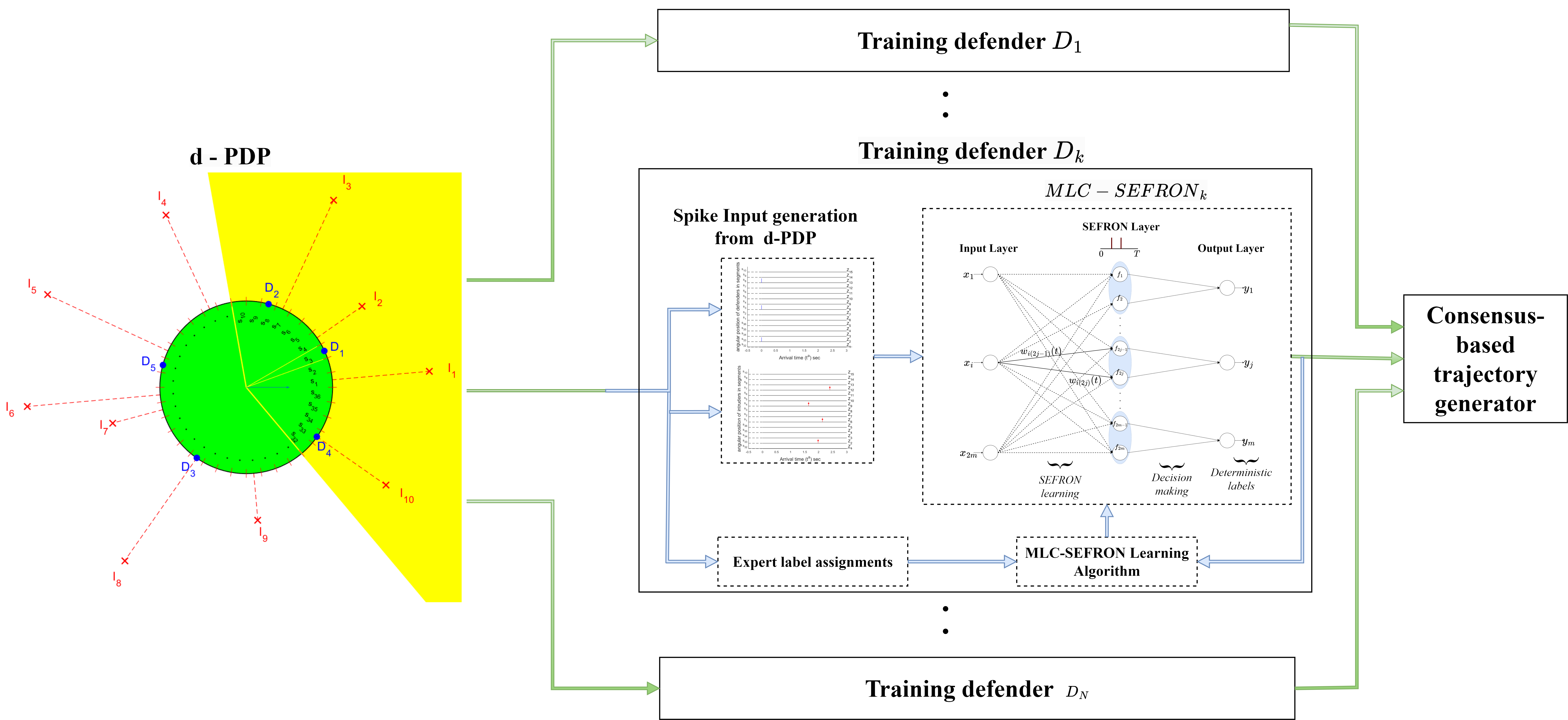}  
\caption{Decentralized spike-based Learning (DSL) framework for discrete perimeter defense problem}.
\label{fig:framework}
\end{figure*}

\section{Related works }
This section presents the related works on key techniques used in this DSL-based solution for d-PDP, namely multi-label learning, and SNNs.

\subsection{Multi-label learning}
In a multi-label learning problem, each sample can have associations with more than one class. The goal of the learning algorithm is to predict all the classes that a given sample is associated with. 
Some existing multi-classification learning algorithms are also used for handling multi-label learning problems.
One of the first approaches for multi-label learning is performed for text categorization \cite{schapire2000boostexter}. In \cite{schapire2000boostexter}, the multi-label learning is used to categorize the text in a news article into multiple topics such as politics, society, etc., Then in \cite{boutell2004mllBR1}, the prediction of each label is performed using a binary classifier. Therefore multiple binary classifiers are designed for learning multiple labels, this strategy is termed as binary relevance strategy. 

Advances in the field of deep learning have led to the usage of deep neural networks for multi-label learning \cite{minling2006rankbasedmll,huang2013sigmoidalmll,zhang2009mimlrbf}. In \cite{minling2006rankbasedmll} and \cite{huang2013sigmoidalmll}, a backpropagation approach is proposed for multi-label learning.  
In \cite{zhang2009mimlrbf}, a clustering algorithm has been used for multi-label learning
Multi-label learning using deep neural networks requires a large amount of data for efficient performance. Moreover, the architecture of deep neural networks consists of many hidden which makes it computationally expensive. Also, each label is predicted in a probabilistic manner, which decreases confidence in the prediction.  Therefore in the proposed DSL framework, deterministic multi-label learning using a binary relevance strategy is performed using an SNN with no hidden layers.  The relevant literature on SNNs is presented in the next section.
\ref{sec:snnrelated works}.

\subsection{Spiking neural networks}
\label{sec:snnrelated works}
Spikes emitted by spiking neurons efficiently embed the spatio-temporal information present in the input given to them \cite{kasabov2014neucube}. The detailed discussion on various existing learning algorithms for SNNs is presented in \cite{dora2021review}. Bohte $et \ al.$ \cite{bohte2002error} developed a gradient-based weight update strategy for SNNs and it is termed as SpikeProp. In SpikeProp, linearities are assumed in the membrane potential of a neuron to evaluate the gradient of a spike. This is done as the gradient of spikes cannot be evaluated directly due to their discontinuous nature. In\cite{gutig2006tempotron}, a supervised learning algorithm for weights in SNN is developed by exploiting spatio-temporal features. Spike Time Dependent Plasticity (STDP) \cite{markram1997regulation}, is another learning technique developed for SNNs that is motivated by biological mechanisms. In rank-order learning for SNNs described in \cite{delorme2001rol2} and \cite{wang2015spiketemprol1}, the weight updates are evaluated based on the spiking rate of the neuron. To increase the performance further evolving layers are used in SNN architecture by neuron addition or deletion strategies as in \cite{dora2016sresn, dora2018tmmsnn}.

In \cite{abee2019sefron}, SNN architecture with time-varying weights is proposed for classification. The SNN in \cite{abee2019sefron} is termed as Synaptic Efficacy Function based leaky-integrate and fire neuRON (SEFRON)\cite{abee2019sefron}. A novel learning algorithm for weight updates is proposed in SEFRON which distributes the evaluated weight updates over time. 
The SEFRON results clearly highlight that the proposed synapse model, with its time-varying nature, executes binary classification tasks with high computational power in comparison to other SNN learning algorithms now in practice. In \cite{mc-sefron} the SEFRON is extended for multi-class classification problems. The results in \cite{abee2019sefron} and \cite{mc-sefron} point out that modeling weights with time-varying functions in SNN increase the classification performance of the network. 
The aforementioned SNN learning algorithm requires additional encoding mechanisms to convert the real-valued data into spikes. Whereas the STMTA problem posed by d-PDP has inherent spatio-temporal nature. Hence STMTA problem can be directly solved using SNNs without any extra encoding mechanisms. A Multi-Label Classifier using SEFRON (MLC-SEFRON) network is proposed in this paper to handle multi-task assignment problems. This is the first time in the literature that an SNN is used to solve a multi-task assignment problem. 
 
\section{Mathematical formulation of d-PDP as Decentralized Spiking multi-label learning problem} \label{sec:problem_formulation}	
The mathematical description of discrete PDP (d-PDP) as a spiking multi-label learning problem is presented initially in this section. Figure \ref{fig:framework} shows the proposed Decentralized Spike-based Learning (DSL) framework. The spatio-temporal location information of defenders and intruders is represented as spikes and given as input to the MLC-SEFRON network.
The solution of STMTA is used as an expert for the training of the network to learn assignments as the labels. The proposed DSL framework employs the decentralized approach, in which a defender learns its assignments based on the  spatio-temporal inputs from its own perspective. 
The labels of assigned locations to a defender are learned using the MLC-SEFRON learning algorithm. The predicted labels are then post-processed to compute the conflict-free assignments for all the defenders.  This section also describes the generation of defender trajectory  using the predictions of the MLC-SEFRON network.

\subsection{Discrete Perimeter Defense Problem (d-PDP)}
This paper considers the multi-player perimeter defense problem in which a team of defenders protects a given territory from invading intruders. It is assumed that intruders are moving radially inwards with a constant velocity, as proposed in \cite{macharet2020adaptive,adler2022role}. Defenders are restricted to operate only on the perimeter \cite{shishika2018local,shishika2020cooperative}. Each defender's trajectory is computed cooperatively such that it can capture the intruders assigned to it in a cost-effective sequence.

Without loss of generality, we assume a circular territory $\Omega$ of unit radius centered at origin $O$ whose perimeter $\partial \Omega$ is given as
\begin{align}
\Omega &= \left\{ \bm{p} \in \Re^2  \  \Big| \    {\|\bm{p}  \|}_2 \le 1 \right\}, \\ 
\partial \Omega &= \left\{ \bm{p} \in \Re^2  \  \Big|   {\|\bm{p} \|}_2 = 1 \right\}    
\end{align}
 where $\bm{p}$ is a point in 2-dimensional plane.
This paper considers discrete PDP in which the perimeter is divided into $N_s$ segments  $ (s_1, ..., s_{N_s})$, of equal arcs  as shown in Figure \ref{fig:pdp}. 
 
Let us consider $N$ defenders $\{D_1, \cdots, D_i, \cdots, D_N\}$ operating on the perimeter. The initial position of the defender $D_i$ on the perimeter is given in by $p_i^D = (r, S^D_i  )$. $r$ represents the distance from the origin ($O$) and is always equal to $1$ as defenders operate only on the perimeter. $S^D_i $ represents the segment of defender $D_i$. The motions of the defenders are limited by a maximum angular velocity $v^D$.

Consider $M > N$ intruders  $I_1, \cdots, I_j, \cdots,I_M$ radially moving towards the perimeter with speed $v^I$ . The position $p^I_j$ of the $j^{th}$ intruder is represented by $p_j^I =(r_j^I,S^I_j )$. $r_j^I$ represents the distance of the intruder $I_j$ from the center of the territory. $S^I_j$ represents the segment of intruder $I_j$. Note that the positions of the intruders are initialized outside the territory, (i.e., $r_j^I > 1$). The kinematic equation of intruder $I_j$ is given as,  $   - \dot{r}_j^I = v^I$.

 If an intruder crosses the perimeter from any of the segments, then it is considered that the defenders have failed to defend the perimeter. If a defender is present in a given segment when an intruder enters that segment then the defender is considered to have captured the intruder. If $S^D_i $ and $S^I_j $ represent the segment locations of defender $D_i$ and intruder $I_j$ respectively, then capture of the latter is written as
 \begin{align} 
Q(I_j) = \left\{  D_i \Big|  r_j^I(t) =  1  \  \&  \  S^I_j(t)) =  S^D_i(t))   \right\} 
\end{align}
\begin{figure}[hb!]
\centering
\includegraphics[width=0.8\linewidth]{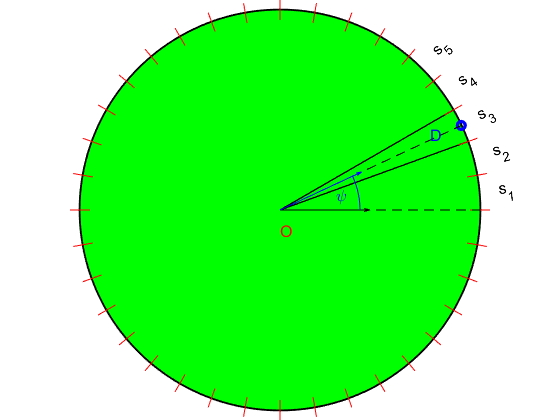}  
\caption{Discrete Perimeter defense problem. The center of the perimeter is  shown with \textcolor{red}{O}, and segments on the perimeter are denoted by $ s_1,s_2,s_3,s_4,s_5$, etc, \textcolor{blue}{D} denotes the defender and $\psi$ denotes it's angular position of on the perimeter respectively}. \
\label{fig:pdp}
\end{figure}
\begin{figure*}[htbp!]
\centering
\subfloat[]{\includegraphics[width=0.27\textwidth]{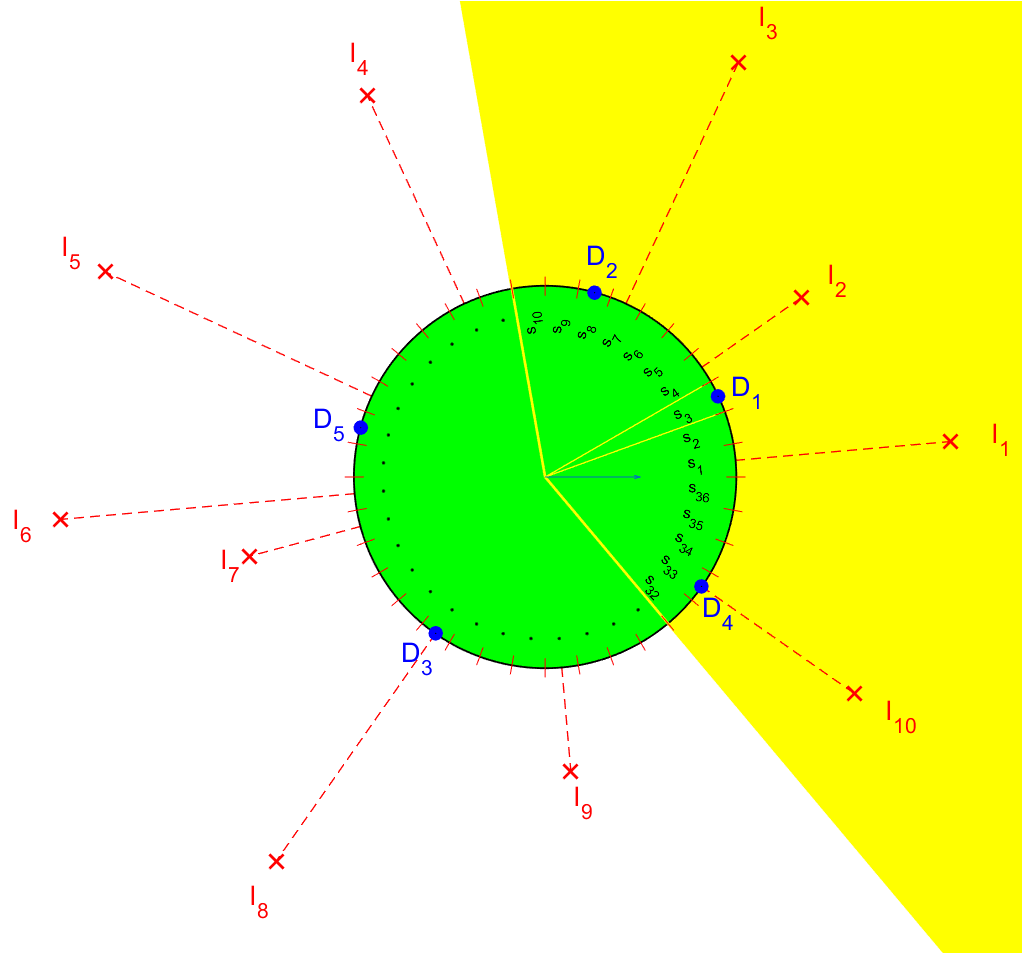}%
\label{fig:pdp_zones}}
\hfil
\subfloat[]{\includegraphics[width=0.33\textwidth]{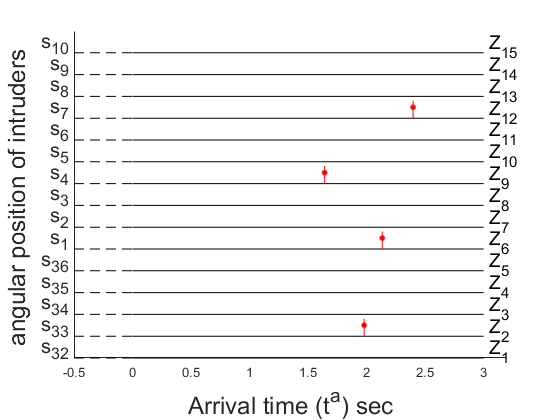}%
\label{fig:input_intru}}
\hfil
\subfloat[]{\includegraphics[width=0.33\textwidth]{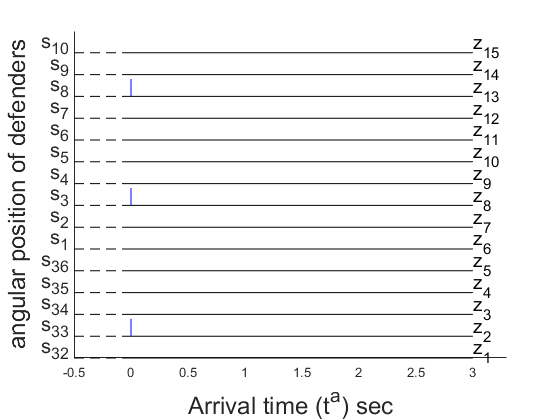}%
\label{fig:input_def}}
\caption{Spike-based representation of defenders and intruders. Figure \ref{fig:pdp_zones} shows the defender locations as $\color{blue}{\bullet}$ and intruder locations as $\color{red}{\times}$. The partial observation range of defender present in segment $s_3$ is shown by the shaded yellow region. Figure \ref{fig:input_intru} and Figure \ref{fig:input_def} shows the spike representation of the intruders  and defenders present in shaded yellow region. }
\label{fig:SNN_PDP}
\end{figure*}

Based on the heading and speed of the intruder $I_j$, it is possible to estimate it's arrival time $t_j^a$ and location $\bm{p}^T_j$ on the perimeter.  To defend the perimeter, a defender has to perform the spatio-temporal task of intercepting the intruder $I_j$ at position $\bm{p}_j^T$ and at time $t_j^a$.  Each defender needs to capture multiple intruders as there are more intruders than defenders. Thus, the team of defenders needs to solve a spatio-temporal multi-task assignment (STMTA) problem \cite{velhal2022decentralized, velhal2022DREAM}.

\subsection{Spike-based representation  for d-PDP}
\label{sec:stmtatosnn}
The technique developed in this section utilizes the capabilities of SNNs to represent both spatial and temporal information relevant for d-PDP. 
Consider the perimeter has been divided into n segments, each of which makes an angle of $(360/n )^{\circ}$ at the center. These segments are denoted as $ \{ s_1, s_2, \cdots,s_i, \cdots,s_n \}$. Each defender has observability of $m$ segments, where $ m \le n$. 
A defender in segment $s_i$, can  observe m segments i.e.
$ \left\{  s_{i + \ceil*{\frac{m-1}{2} } } , \cdots, s_{i-1}, s_i,s_{i+1}, \cdots, s_{i + \floor*{ \frac{m-1}{2} }  }   \right\} $.  (All the negative and zero indices of segments  are converted to positive indices in a cyclic fashion. )
For the decentralized setting, the zones are defined from the perspective of each defender to represent the information. The zones for defender present in segment $s_i$       \big( where $s_i   \Leftrightarrow  z_{\ceil*{ \frac{m+1}{2} }}$ \big)
 are  $ \left\{ z_m, z_{m-1}, \cdots, z_{ \ceil*{ \frac{m+1}{2} } }, \cdots, z_2,z_1  \right\} $.

Figure \ref{fig:pdp_zones} shows a scenario with  5 defenders and 10 intruders for 36 segments and observation of 15 zones. The perimeter has been divided into 36 segments each of which makes an angle of $10^{\circ}$ at the center. These 36 segments are denoted as $ \{ s_1, s_2, \cdots,s_{36} \}$. The current positions of intruders and defenders have been shown using $\color{red}{\times}$ and $\color{blue}{\bullet}$, respectively. The radial trajectory of the intruders towards a specific segment on the perimeter is shown using red dashed lines.

For the partial observation scenario considered in this paper, it is assumed that defenders have a limited sensing range of $150^{\circ}$, which constitutes 15 segments.
For example, the yellow region in the figure represents the sensing range for $D_1$ which implies that it can perceive the spatio-temporal information pertaining of both, intruders and defenders. The segments $\{s_{10}, s_9, \ldots, s_{33}, s_{32} \}$ segments are referred as zones of that defender ($D_1$), represented by  ($z_{15}, z_{14}, \ldots, z_2, z_1$) as shown in  Figures \ref{fig:input_intru} \& \ref{fig:input_def}.  Similar to segments, the zones are also numbered in the anticlockwise direction from $z_1$ to $z_{15}$.
Defender is always assumed to be present in $z_8$, which is the central zone in its sensed region. The zones contain the decentralized information sensed by each defender, on the contrary, segments contain global information. In other words, segments of the territory are designed beforehand and do not depend on the positions of defenders. 

The spatio-temporal information pertaining to intruders and defenders present in the zones of a defender $D_1$ is represented using  spike patterns and shown in Figure \ref{fig:input_intru} and \ref{fig:input_def}. These spikes are event-triggered based on the distance of the object (either intruder or defender) from the perimeter.
The time of a given spike in Figure \ref{fig:input_intru} is directly proportional to the arrival time of the intruder in the corresponding zone. For instance in Figure \ref{fig:input_intru}, the time of arrival of intruder $I_1$ in zone $z_6$ at the perimeter is shown using a spike at $2.15s$. Similarly, a defender in zone $z_{13}$ (i.e. segment $s_8$) is represented using a spike at $0s$.The position of the defender whose perspective is being considered is always considered at the $8^{th}$ zone (middle zone) in 15 segments observation scenario (partial observation scenario). The time of the spikes that represent the position of defenders is always set to $0s$ as defenders are constrained to move along the perimeter. The spike patterns generated are directly presented as input to the MLC-SEFRON via the input layer without any extra pre-processing step.

\subsection{Multi-task assignments to multi-label learning}
 The solution of spatio-temporal multi-task assignments gives the assignments of defenders to the intruders. 
In a decentralized setting of defender $D_i$, each assigned intruder can be identified by the corresponding zone.  Using the assignment solution, one can label the assignment of defenders to the zones. For the labels of defender $D_i$, if it is assigned to a zone $z_j$, then $z_j$ is labeled as $TRUE$. A defender can have multiple assignments, all the assigned segments are labeled as $TRUE$.  The assignment solution provides the target labels ($T^l$) for all the observed m zones  ( $T^l \in {\{1,0\}}^m $). In this way, the assignments of defender $D_i$ are converted into multiple deterministic labels for zones of $D_i$. 
One should note that the centralized solution to the STMTA problem provides the assignments of all the defenders. In the DSL framework assignments of each defender's assignments (with input from that defender's perspective)  are used to train the individual MLC-SEFRON network. In this way, training is scalable and independent of specific defenders.
The spiking network needs to learn these labels. The detailed learning algorithm is discussed in section \ref{sec:architecture}.

\subsection{Trajectory generation from MLC-SEFRON output}
\label{sec:trajectory_gen}
The suitable trajectory for the team of defenders is generated using zones assigned by the MLC-SEFRON to each defender. Predicting assigned zones for each defender in a decentralized manner can lead to a situation where multiple defenders are assigned to a single zone. Also during SNN prediction, there is a chance of assigning a defender to a zone where no intruder is present. To resolve these two issues, a consensus-based bundle algorithm (CBBA) \cite{choi2009consensus} is used to compute the final task assignments.

Lets consider that $\hat{l}_j$ represents the prediction for the $j^{th}$ segments' assignment to a defender. To exploit the spatial correlation between neighboring segments, the prediction for the $j^{th}$ segment is updated using the predictions for the segments in the neighborhood. Further, the final assignment for a segment is assigned to $0$ if there is no intruder present in that segment. Based on this, the effective predictions ($\hat{l}^{\text{eff}}_j$) for a given segment are given by
\begin{align}
\hat{l}^{\text{eff}}_j = \begin{cases}
( \hat{l}_j  + \alpha*  \hat{l}_{j+1} + \alpha*  \hat{l}_{j-1}), & {\text{Intruder present in $s_j$}} \\
0,  & {\text{Intruder absent from $s_j$}}
\end{cases}
\end{align} 

where  $ \alpha \in  [  0, 1 ] $ is the scaling factor that governs the impact of predictions for neighboring segments on a given segment. If $ \alpha = 0$, there is no effect of predictions of neighboring segments.

Every defender computes its own trajectory by arranging the intruders present in the assigned zones in ascending order of their arrival times. Every defender computes the trajectory based on their individual effective labels ( $\hat{l}^{\text{eff}}_j $ ). It is possible that multiple defenders are assigned a given segment because the predictions for each defender are generated independently. To determine final assignments, a distance-based cost function is used. Among these defenders, each one (say $D_i$) bids for the $s_j$ segment with a  cost $l_{ij}$, where $l_{ij}$ is the effective distance $D_i$ needs to travel from its current location to  $j^{th}$ segment. 
\begin{align}
l_{ij} = \hat{l}^{\text{eff}}_j * arc(S'^D_i),s_j)
\end{align} 

where $S'^D_i$ is the segment location from which defender $D_i$ has to start in order to capture the intruder in the $s_j$ segment. To fix the issue of multiple defender assignments to a single $s_j$ segment, the defender which has the minimum cost is considered the winner in the bidding. Based on this bidding and consensus algorithm final trajectory for each defender is generated.

\section{MLC-SEFRON Architecture \&  Learning Algorithm} \label{sec:architecture}
In this section a Multi-Label Classifier using SEFRON (MLC-SEFRON) to predict the labels of multiple zones assigned to a single defender in a deterministic manner is presented. The MLC-SEFRON architecture is described first, followed by its learning algorithm.
\begin{figure}[h]
\centering
\includegraphics[width=0.9\columnwidth]{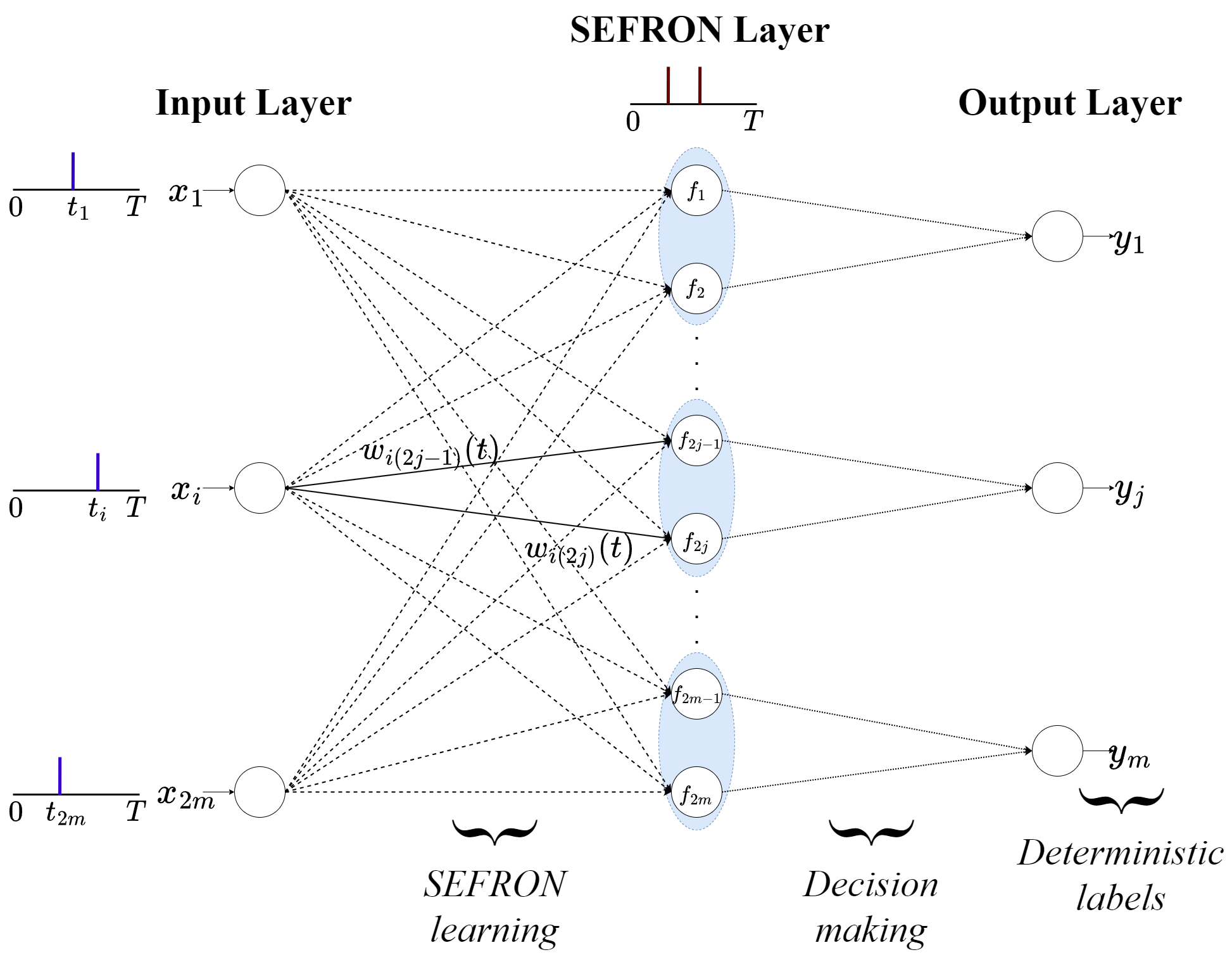}
\caption{{Architecture of MLC-SEFRON}}.
\label{fig:architecture}
\end{figure}
\subsection{Architecture of MLC-SEFRON}
The architecture of MLC-SEFRON for training a defender with a sensing range having $m$ zones is shown in Figure \ref{fig:architecture}. The input layer of MLC-SEFRON consists of $2m$ neurons which are used to present the spike patterns illustrated in Figures  \ref{fig:input_def} \& \ref{fig:input_intru} (which  represent the spatio-temporal information about defenders and intruders). The first $m$ neurons are used to represent the defender information, whereas the last $m$ neurons are used for representing intruder information. Let $x_i = \{ t_i \}$  be  the input spike pattern presented through the $i^{th}$ input neuron, where $t_i$ denotes the time of spike of that neuron. The SEFRON layer in the architecture also consists of $2m$ spiking neurons ($f_1, ..., f_{2m}$).  The weight of the synapse between $i^{th}$ input neuron and ${2j}^{th}$ neuron in the SEFRON layer is denoted as $w_{i2j}(t)$. Each weight $w(t)$ connecting the input layer and SEFRON layer is modeled as a time-varying Gaussian function as described in SEFRON \cite{mc-sefron}.  The output layer consists of $m$ neurons associated with $m$ zones in the perimeter. The response of $j^{th}$  output neuron depends on the spike responses of ${2j-1}^{th}$ and ${2j}^{th}$ neurons in the SEFRON layer as shown below
\begin{align}
        y_j=\begin{cases}
        1, & {\text{If } \hat{t}_{2j-1} < \hat{t}_{2j}} \\
        0, &  {\text{If } \hat{t}_{2j-1} \geq \hat{t}_{2j}}
    \end{cases}
\label{eq:label}
\end{align}

where $\hat{t}_{2j-1}$ and $\hat{t}_{2j}$ represents the time of first spikes of ${2j-1}^{th}$ and ${2j}^{th}$ neurons in the SEFRON layer. $y_j$ represents the deterministic label predicted for the $j^{th}$ zone in the perimeter. If $y_j$ is $1$ then a defender $D$ is assigned to $j^{th}$ zone and vice versa if  $y_j$ is $0$. In a similar fashion, labels are evaluated for all $m$ neurons in the output layer. If $\hat{C}$ denotes the predicted multi-label output from the MLC-SEFRON network then it is given as
\begin{align}
        \hat{C} = \{\hat{c}_1,...\hat{c}_j,...\hat{c}_m\},  \text{where  } \hat{c}_j = y_j
        \label{eq:multi-label output}
\end{align}
If $\hat{t}_{j}$ denotes the spike time ${j}^{th}$ neuron in the SEFRON layer then its spike time can be evaluated as
\begin{align}
        \hat{t}_j= \{t|v_j(t)=\theta_j\} 
\end{align}
where $\theta_j$ and $v_{j}(t)$ denote the potential threshold and potential of ${j}^{th}$ neuron in the SEFRON layer respectively. 
The potential thresholds for all neurons in the SEFRON layer are initialized as described in \cite{mc-sefron}.
The potential $v_{j}(t)$ is evaluated as
\begin{align}
        v_j(t) = \sum_{i=1}^{2m} w_{ij}(t_i)  * \epsilon (t-t_i)
        \label{eq:potential}
\end{align}

where $\epsilon (t-t_i)$ is the unweighted membrane potential induced at time $t$ by input spike at $t_i$. It is modeled using the spike response function \cite{bohte2002error}, given as
\begin{align}
        \epsilon(s) = \frac{s}{\tau} \exp({1- \frac{s}{\tau}})
\end{align}
where $\tau$ is the time constant of the neuron.

The actual assignment prediction $\hat{c}_j$ is then compared with expert assignment $c_j$. The expert assignments for each zone are  evaluated using an expert solution (see Section \ref{sec:data generation}). In the next section, an MLC-SEFRON learning algorithm is described.

 \subsection{MLC-SEFRON learning algorithm}
 \label{sec: learning}

MLC-SEFRON uses three strategies for learning, namely \textit{initialization strategy}, \textit{escaped intruder strategy} and \textit{incorrect assignment strategy}. In the \textit{initialization strategy} the potential thresholds of the neurons in the SEFRON layer and weights connected to them are initialized as described in SEFRON \cite{mc-sefron}.
In \textit{escaped intruder strategy} weights are updated such that the defender is assigned to the zone. In  \textit{incorrect assignment strategy}  weights are updated such that the defender is not assigned to the zone. Next, these three different strategies are explained in detail. 

\subsubsection{Initialization strategy}
\label{sec:initiali}

Let us consider the initialization of weights connecting to ${2j}^{th}$ neuron in the SEFRON layer and its threshold $\theta_{2j}$. The first input sample from the dataset which has a desired prediction that a defender should not be assigned to $j^{th}$ zone is used for this purpose. This initialization is done as shown below
\begin{align}
    w_{i(2j)}(t) &= u_{i(2j)}(T_d) \exp(-\frac{(t-t_i)^2}{2\sigma^2})
     \nonumber\\
    \theta_{2j} &= \sum_{i=1}^{2m} u_{i(2j)}(T_d) \epsilon (T_d-t_i) \quad  \forall {j \in \{1,\cdots, m\}}
    \label{eq:th_init}
\end{align}
where $T_d$ is defined as the ideal firing time $T_d$ is chosen such that a neuron in the SEFRON layer can utilize the information present in incoming spike patterns to make a decision. If $T_d$ is chosen close to the start of the simulation, then MLC-SEFRON is not able to utilize the information present in spike patterns. If $T_d$ is chosen close to the end of the simulation (i.e., $T$), then MLC-SEFRON takes a longer time to make a decision. Which leads to the inaccurate assignment prediction for zone $z_j$ by MLC-SEFRON. $T_d$ is chosen appropriately as mentioned in \cite{mc-sefron}.

In Equation (\ref{eq:th_init}), $u_{i(2j)}(T_d)$ is defined as the fractional contribution of $i^{th}$ input neuron for ${2j}^{th}$ neuron in the SEFRON layer to spike at $T_d$.
\begin{equation}
        u_{i(2j)}(T_d)= \frac{\delta w (T_d-t_i)}{\sum_{i=1}^{2m}\delta w (T_d-t_i)}
\end{equation}
where $\delta w (.)$ is defined as the STDP weight update \cite{markram1997regulation}. It is computed as
\begin{equation}
\delta w (s)= \begin{cases}
A_+ \exp(-\frac{s}{\tau_+}), \text {if} \hspace{0.1cm}s\geq0\\
-A_- \exp(\frac{s}{\tau_-}), \text {if} \hspace{0.1cm}s<0
\end{cases}
\end{equation}

where $A_+,A_- $ are the maximum weight changes allowed and $\tau_+,\tau_-$ are the time constants for STDP. 
Similarly for initializing the threshold of ${2j-1}^{th}$ neuron and weights connected to it, the first input sample which has a desired prediction that a defender should be assigned to $j^{th}$ zone is used.
In this manner, the weights connected to all neurons in the SEFRON layer and their thresholds are initialized.
The values of constants such as $T, \sigma, \tau_+$ and $\tau_-$  are set as in \cite{mc-sefron}.

 \subsubsection{Escaped intruder strategy}

This strategy is used when a defender is not assigned to a zone in which an intruder is present. Given by

\textbf {If $\hat{c}_j \neq c_j$ and $\hat{t}_{2j-1} \geq \hat{t}_{2j}$}

In this case, the misclassifcation occurs because ${2j}^{th}$ neuron spikes earlier than ${2j-1}^{th}$ neuron in the SEFRON layer respectively. To fix this issue the weights connected to these neurons are updated such that they spike at desired firing times denoted by ${t^d_{2j-1}}$ and ${t^d_{2j}}$ respectively,  which are given by
\begin{align}
    t^d_{2j-1} &= \begin{cases}
     \text{$\hat{t}_{2j-1}$},  & {\text{if} \ \hat{t}_{2j-1} < T_d}\\
     \text{$T_d$},   & {\text{if} \  \hat{t}_{2j-1} \geq T_d}
     \end{cases} \nonumber\\
    t^d_{2j} &= \begin{cases}
     \text{$\hat{t_{2j}}$}, & {\text{if} \ \hat{t_{2j}}\geq (t^d_{2j-1} + T_m)}\\
     \text{$t^d_{2j-1} + T_m$}, &{ \text{if} \ \hat{t_{2j}} < (t^d_{2j-1} + T_m)}
     \end{cases} 
     \label{eq:d1}
\end{align}

where $T_m$ is defined as the margin threshold required between two neurons in the SEFRON layer for better generalization. If $T_m$ is set very small value (i.e., $0$), then MLC-SEFRON is not able to classify the highly overlapped samples.  Also, if $T_m$ is set to a high value then MLC-SEFRON takes a longer time to converge. $T_m$ is chosen appropriately as described in \cite{mc-sefron}.

Then weight update strategy as described in \cite{mc-sefron} is used to update the weights connected to ${2j-1}^{th}$ and ${2j}^{th}$ neurons in the SEFRON layer such that they spike at ${t^d_{2j-1}}$ and ${t^d_{2j}}$ respectively. 

Without loss of generality the weight update for $k^{th}$ neuron  in the SEFRON layer with a desired firing time $t^d_k$ is described further. 
The error function $e$ to evaluate the update is given as
\begin{align}
        e= \frac{\theta_k}{\hat{V_{k}}(t^d_k)} - \frac{\theta_k}{\hat{V_{k}}(\hat{t_k})}
    \label{eq:error}
\end{align}
The error function $e$ is designed as above because it can be  multiplied directly with the fractional contribution $u$  to get the required update value. If $\Delta w_{ik}$ denotes the required weight update then it is given as
\begin{align}
     \Delta w_{ik}=  \lambda \hspace{0.1cm} u_{ik}(t^d_k)\hspace{0.1cm} e 
    \label{eq:update}
\end{align}
where $\lambda$ is the learning rate. In Equation (\ref{eq:error}), $\hat{V_{k}}(\hat{t})$ is defined as the potential required for $k^{th}$ neuron in the SEFRON layer to  spike at $\hat{t}$.
\begin{align}
       \hat{V_{k}}(\hat{t})= \sum_{i=1}^{2m} u_{ik}(\hat{t})\hspace{0.1cm} \epsilon (\hat{t}-t_i)
\end{align}

The update $\Delta w_{ik}$ is modulated using a gaussian function and then the existing weight is updated as given below

\begin{align}
     w_{ik}(t) \leftarrow w_{ik}(t) + (\Delta w_{ik}\exp(-\frac{(t-t_i)^2}{2\sigma^2}))
    \label{eq:fin_update}
\end{align}
   
The weights of ${2j-1}^{th}$ and ${2j}^{th}$ neurons are updated using Equations \ref{eq:error} to \ref{eq:fin_update} to resolve the misclassification.

\subsubsection{Incorrect assignment strategy}

This strategy is used when a defender is wrongly assigned to a zone. Given by

\textbf {If $\hat{c}_j \neq c_j$ and $\hat{t}_{2j-1} < \hat{t}_{2j}$}

In this case, the misclassifcation occurs because ${2j-1}^{th}$ neuron spikes earlier than ${2j}^{th}$ neuron  in the SEFRON layer. To fix this issue the ${t^d_{2j-1}}$ and ${t^d_{2j}}$  are evaluated as
\begin{align}
    t^d_{2j} &= \begin{cases}
     \text{$\hat{t_{2j}}$}, & {\text{if} \  \hat{t_{2j}} < T_d}\\
     \text{$T_d$}, &{ \text{if} \ \hat{t_{2j}} \geq T_d}
     \end{cases} \nonumber\\
         t^d_{2j-1} &= \begin{cases}
     \text{$\hat{t}_{2j-1}$},  & {\text{if}  \ \hat{t}_{2j-1} \geq (t^d_{2j} + T_m)}\\
     \text{$t^d_{2j} + T_m$},   & {\text{if} \ \hat{t}_{2j-1} < (t^d_{2j} + T_m)}
     \end{cases} 
     \label{eq:d2}
\end{align}

To make actual assignment $\hat{c}_j$ similar to that of desired assignment $c_j$ the weights connected to  ${2j-1}^{th}$ and ${2j}^{th}$ neurons in the SEFRON layer respectively are updated using Equations \ref{eq:error} to \ref{eq:fin_update}.

To summarize the MLC-SEFRON learning algorithm is used to predict the labels of multiple segments assigned to a defender in a deterministic manner. The obtained labels of segments that are assigned to a defender are used for its trajectory generation.

\section{Simulation results} \label{sec:result}
This section presents the results of performance evaluation and comparison of the proposed DSL framework with the existing STMTA formulation from \cite{velhal2022DREAM}. Initially in this section, the dataset is generated using the expert (STMTA) approach for the proposed DSL framework of d-PDP is discussed. Then the performance of the proposed MLC-SEFRON for training a single defender is studied using multi-label metrics such as $Precision$, $Recall$, and $F1-score$ \cite{mll-metrics1}. Finally, the success percentage achieved by the team of defenders in capturing the intruders is evaluated and compared with the expert approach. The simulations are carried out using Matlab 2022a, in a Windows 10 system with 16GB memory, 6 cores, and a 3.2GHz machine. 

\subsection{Expert policy-based data generation} 
\label{sec:data generation}
A circular perimeter is divided into 36 segments. Two different scenarios are considered for generating a dataset one is a full observation scenario in which a defender senses all 36 segments. The other is a partial observation scenario, in which a defender senses  15 segments of the perimeter.
 The dataset is generated  with a team of 5 defenders, placing them randomly along the perimeter. The intruders are spawned using Poisson distribution with an arrival rate of 4 over a period of 8 seconds. The intruders are placed randomly using a uniform distribution over all segments. The dataset is generated using random Monte Carlo runs, with at most one intruder spawned in a segment in each run.  The d-PDP is formulated as the spatio-temporal multi-task assignment (STMTA) problem as described in \cite{velhal2022decentralized,velhal2022DREAM}.  
 This STMTA problem is solved by an expert policy similar to  approaches presented in\cite{velhal2022DREAM,chopra2014heterogeneous}. The detailed steps of expert policy are given below.  The expert policy computes the assignments by minimizing the distance traveled by the defenders along the perimeter while maximizing the number of captures. 
The cost function and the optimization problem used by  an expert policy  are discussed below.
   
 \subsubsection{Cost function} 
 The defender needs to capture the intruder either as a first task from its own location or as a subsequent task after capturing another intruder. The main difference is the starting point for the execution of the task. For the first task, the defender will start from its own location, and for subsequent tasks, it starts from  the previously captured intruder. The cost is defined as the arc distance that needs to be traveled by the defender along the perimeter to capture an intruder. If this traveling distance is infeasible due to the velocity limit, the cost is selected as the large value ( denoted by $\kappa$) to (avoid or at least) minimize  such assignments.  If it requires traveling in negative time,  the cost is selected as infinite to avoid assignments.  Here, $ d(\arc{AB})$  denotes arclength along the arc AB.
Mathematically cost function is given as, 
\begin{align}
 &c^f_{ij} = \begin{cases}   d(\arc{S_i^D,S^I_j} ) & \text{if } \dfrac{ d(\arc{S_i^D,S^I_j} )}{V_{max}^D} \le t^a_j \\ 
\kappa & \text{ otherwise}
\end{cases} \label{chap3_eq:cost_first}
\end{align}
\begin{align}
 & c^s_{kj} = \begin{cases}   d(\arc{S_k^T,S^I_j} ) & \text{if } \dfrac{ d(\arc{S_k^T,S^I_j} )}{V_{max}^D} \le t^a_j - t^a_k \\ 
\kappa & \text{if } \dfrac{ d(\arc{S_k^T,S^I_j} )}{V_{max}^D} > t^a_j - t^a_k \\
  \infty  & \text{if } t^a_j < t^a_k \end{cases} \label{eq:cost_sub} \\ 
  &\qquad \text{for } i \in \mathcal{I} = \{1,2,\cdots,N\} ,\quad j \in \mathcal{J} = \{1,2,\cdots,M \} , \nonumber \\
  &\qquad \qquad k \in \mathcal{K} = \{1,2,\cdots,M-1\}  \nonumber 
\end{align}

\subsubsection{Optimization Problem}
The formulated optimization problem is a linear sum assignment problem where decision variable $\delta^f_{ij} $ is used to decide whether a defender $D_i$  is assigned to task $T_j$ as a first task or not.  Another decision variable  $\delta^s_{kj} $ is used to decide whether a defender will be assigned to task $T_j$ just after the task $T_k$  or not.  Mathematically, the optimization problem is given as 

\begin{subequations} 
\addtocounter{equation}{-1}
\begin{align} \label{chap3_eq:integer_prog}
 \min_{\delta^f_{ij} \ \delta^s_{kj}} & \sum_{i\in \mathcal{I} } \sum_{j \in \mathcal{J} } c^f_{ij} \delta^f_{ij} + \sum_{k\in \mathcal{K} } \sum_{j \in \mathcal{J} } c^s_{kj} \delta^s_{kj} \ \ \\ 
 {\rm s. \ t.} \ 
 & \delta^f_{ij} \in \{0,1\}\qquad \forall (i,j) \in { \mathcal{I} \times {\mathcal{J} } } \label{chap3_eq:cost_cond_1a} \\
 & \delta^s_{kj} \in \{0,1\}\qquad \forall (k,j) \in { \mathcal{K} \times {\mathcal{J} } } \label{chap3_eq:cost_cond_1b} \\
 &\sum_{i \in \mathcal{I}} \delta^f_{ij} + \sum_{k \in {\mathcal{K} } } \delta^s_{kj} = 1 \quad \forall j \in {\mathcal J} \label{chap3_eq:cost_cond_2} \\ 
 &\sum_{j \in {\mathcal{J} }} \delta^f_{ij} \le 1 \quad \forall i \in \mathcal{I} \label{chap3_eq:cost_cond_3a} \\
 & \sum_{j \in {\mathcal{J} }} \delta^s_{kj} \le 1 \quad \forall k \in {\mathcal{K}} \label{chap3_eq:cost_cond_3b} 
\end{align} 
\end{subequations}

The solution to the optimization problem gives the assignments of the defender to intruders. This assignment solution is used as a ground truth for the supervised learning problem. As the expert solution is sub-optimal and it is obtained using a fixed team of defenders, its success rate cannot be 100\% i.e., some of the intruders will penetrate the perimeter. Therefore for generating the dataset, intruders who enter the perimeter are sequentially removed to obtain feasible assignments of remaining intruders.  The sequential removal of these infeasible intruders is not the optimal method \footnote{One needs to solve the combinatorial optimization problem (to decide which tasks should be removed) for computing the optimal solution. Computing the optimal solution is outside the focus of this paper.}. The obtained solution will be sub-optimal, so we call it an expert solution.

\begin{figure*}[t!]
\centering
\subfloat[]{\includegraphics[width=0.48\textwidth]{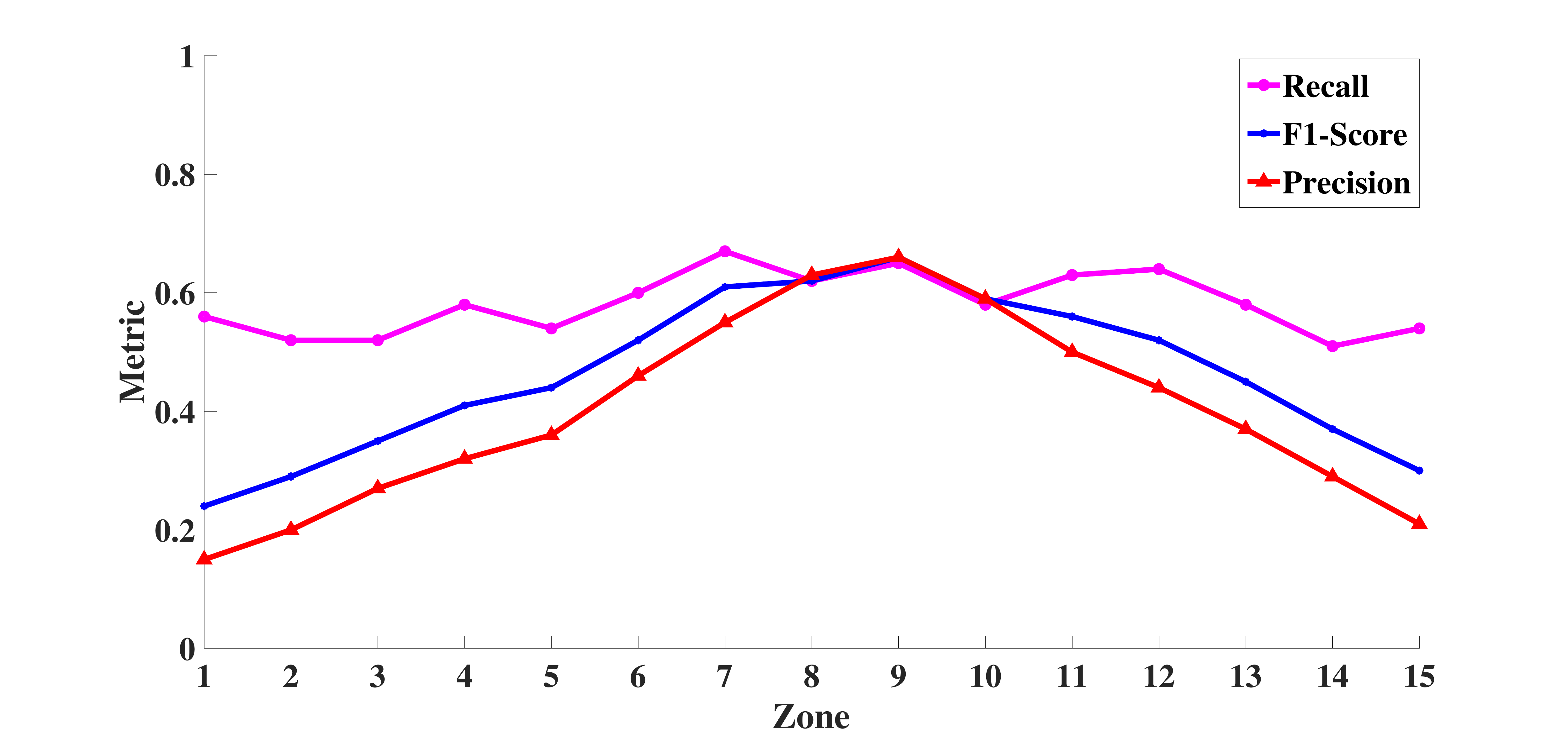}%
\label{fig:MLL_15_segs_results}}
\hfil
\subfloat[]{\includegraphics[width=0.48\textwidth]{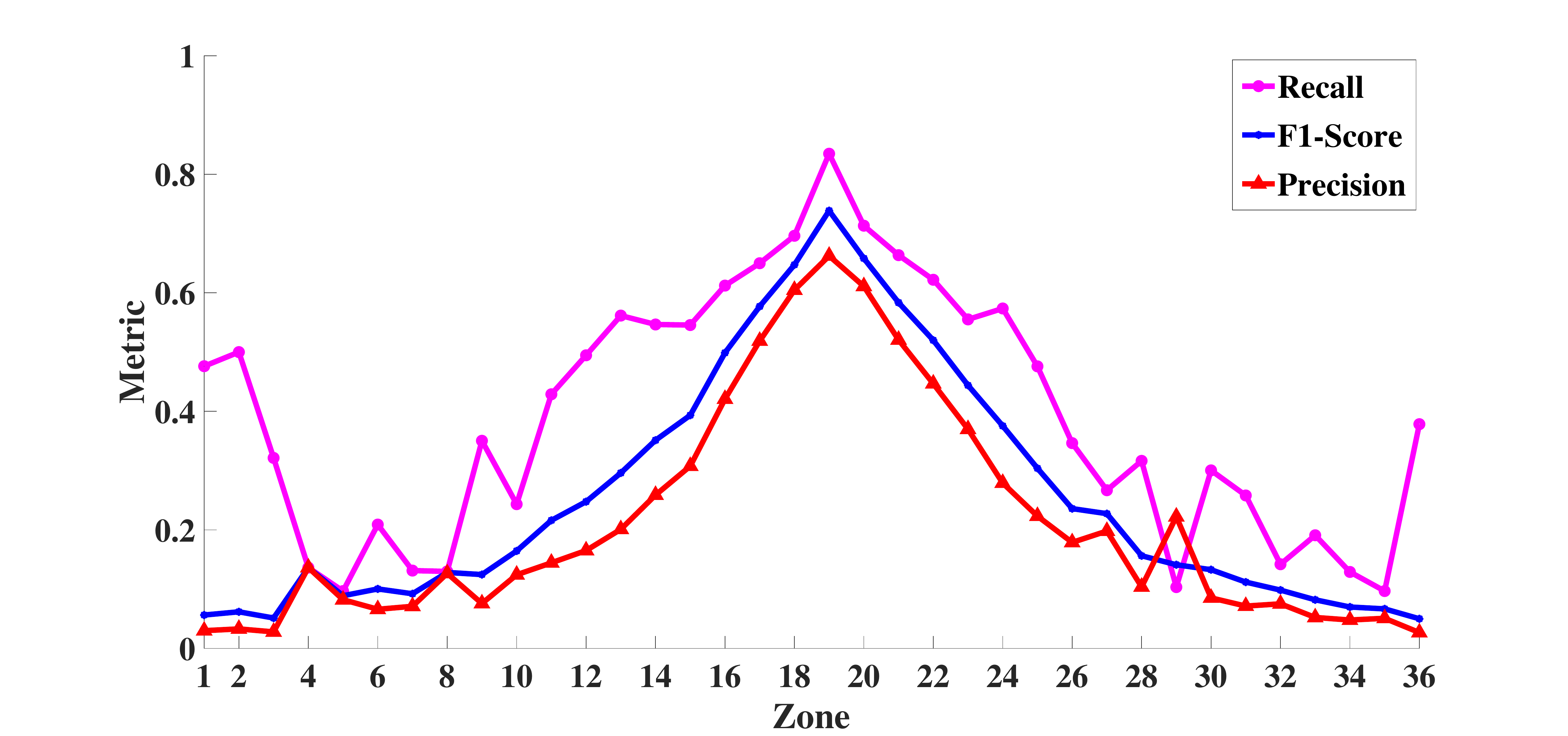}%
\label{fig:MLL_36_segs_results}}
\caption{Multi-label performance evaluation. 
Figure 5a shows the results of performance evaluation of MLC-SEFRON algorithm for a single defender in a partial observation scenario. Whereas Figure 5b shows the results of performance evaluation of MLC-SEFRON algorithm for a single defender in a full observation scenario.  Purple line shows the $Recall$, blue line shows the $Fl-score$ and red line shows the $Precision$ along various zones respectively}
\label{fig:MLL_results}
\end{figure*}

The dataset is generated for each defender in a decentralized fashion. While generating the dataset for a particular defender, the locations of the intruder and  the defender are considered from the perspective of that particular defender. These spatio-temporal representations are carried out as described in section \ref{sec:stmtatosnn}. The desired output for a zone is labeled as assigned if the expert solution assigns a defender to that zone; otherwise, it is labeled as unassigned. If there are more intruders than defenders, the expert (STMTA) solution assigns some defenders to multiple zones.
 
The optimization problem in STMTA formulation of d-PDP  minimizes the total distance traveled by a team of defenders while capturing the maximum number of intruders. Therefore in this solution,  defenders are more likely assigned to the neighboring zones than the further zones to minimize the distance traveled. As the trainee defender is always located in the central zone, the intruders closer to these zones are more likely to be assigned when compared with intruders in the zones farther away from the central zone. 
Technically, the dataset has a long-tail problem, where the number of assigned labels is lower for the further away zones than that of the central zones.

A dataset is generated for a team of five defenders by solving 10000 scenarios of d-PDP using the expert solution. A total of 50000 samples are generated in this dataset, out of which only 10000 are used for training the MLC-SEFRON. The dataset has higher number samples in which trainee defender is assigned to the central zone when compared with the samples assigned to other zones. This is due to the presence of trainee defender in the central zone. Thus the generated dataset using expert policy is unbalanced due to the long-tail problem. To balance this dataset the minority class samples of the dataset are over-sampled using the synthetic minority class over-sampling technique  \cite{chawla2002smote}. This oversampled dataset is used for training. The remaining 40000 samples of the dataset are used for the testing. The results are presented for the testing dataset.

\subsection{Performance evaluation of MLC-SEFRON}

In this section, the results of the performance evaluation of MLC-SEFRON in predicting assigned zones for a single defender are presented. The weights obtained after training a single defender are directly used for evaluating the performance of other defenders. Thus the training in the  DSL framework is decentralized.
Multi-label performance metrics such as $Precision$, $Recall$, and $F1-Score$ are used for this performance evaluation. 
Figure \ref{fig:MLL_results} shows the zone-wise variation of these multi-label metrics for full and partial observational scenarios.
Figure \ref{fig:MLL_15_segs_results} displays the $Precision$, $Recall$, and $F1-Score$ plots evaluated  using the MLC-SEFRON for all 15 zones in partial observation scenario. It can be illustrated that performance measures are greater for central zones and subsequently decline for zones farther from central zones are considered. 
Figure \ref{fig:MLL_36_segs_results} displays the multi-label performance metrics of the MLC-SEFRON for full observation scenario. 
The high $Recall$ performance in the central zone is due to more assigned samples in the dataset for these zones. Whereas the high $Recall$ performance in the farther zones is due to a higher number of un-assigned samples in the dataset for these zones.

\subsection{Performance evaluation of DSL framework for d-PDP} 
\label{sec:performance comparision}
The performance of the team of defenders is evaluated using the success percentage. The success percentage ($S_p$) is defined as
\begin{align}
S_p =\frac{ \text{number of intruders captured}}{\text{total number of intruders}} \times 100
\end{align}

\begin{table}[t!] 
\setlength{\arrayrulewidth}{1.1pt}
\centering
\caption{Success percentage comparisons for STMTA and DSL }
\begin{tabular}{|l|r|r|r|r|}
\hline
            & \begin{tabular}[c]{@{}c@{}}  Expert policy \\(centralized)  \end{tabular}   & \multicolumn{1}{c|}{DSL}  & \begin{tabular}[c]{@{}c@{}} DSL with   \\ neighboring \\ effect \end{tabular} & \begin{tabular}[c]{@{}c@{}} Learning \\ efficiency \\ of DSL\end{tabular}\\ \hline
  \begin{tabular}[c]{@{}c@{}} Full \\ observation \end{tabular}  & 85.0425                    & 74.5195     &     82.8825  & 97.50\\ \hline
\begin{tabular}[c]{@{}c@{}} Partial \\ observation \end{tabular}  & 85.3439                 &  72.5642         &   79.4924 & 93.14  \\ \hline
\end{tabular}
\label{table:comparison_full_info}
\end{table} 

 The table \ref{table:comparison_full_info} shows the success comparison of DSL framework and expert (STMTA) policy.  The STMTA solution is used as a training dataset and hence STMTA is expected to perform better than the DSL framework The success of the STMTA is $84.95\%$, where as SNN gives success of $74.51\%$ for the full observation scenario  i.e. each defender observes all 36 segments and acts accordingly. The performance of the DSL frameworkimproves up to $82.88\%$  by considering the effect of neighbors in post-processing.
Similar results are obtained for the case of partial observations. Here, defenders can sense only  15 segments but the auction of tasks is done in a centralized way. The  STMTA (expert)  policy has a success of $ 85.34\% $ and the DSL framework provides a success of $72.56\%$, and this can be improved to $79.49\% $ by adding the effect of neighbors in post-processing.

The proposed DSL framework with post-processing obtains $97.5\%$ and $93.14\%$  success performance compared to the STMTA (expert) solution for the full and partial observation scenarios, respectively.
\begin{figure*}[b!]
\centering
\subfloat[]{\includegraphics[width=0.45\textwidth]{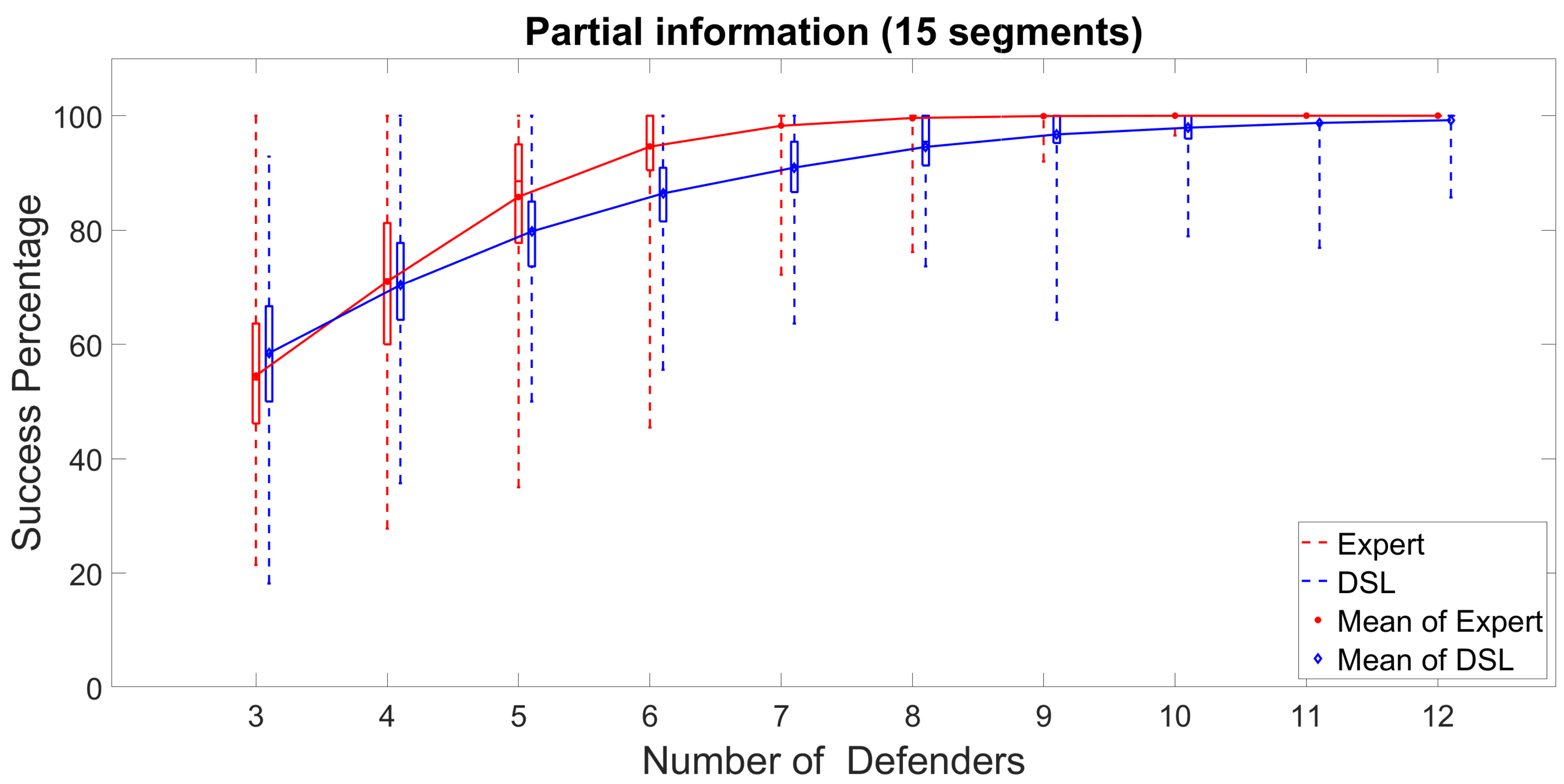}%
\label{fig:scalability_partial}} 
\hfil
\subfloat[]{\includegraphics[width=0.45\textwidth]{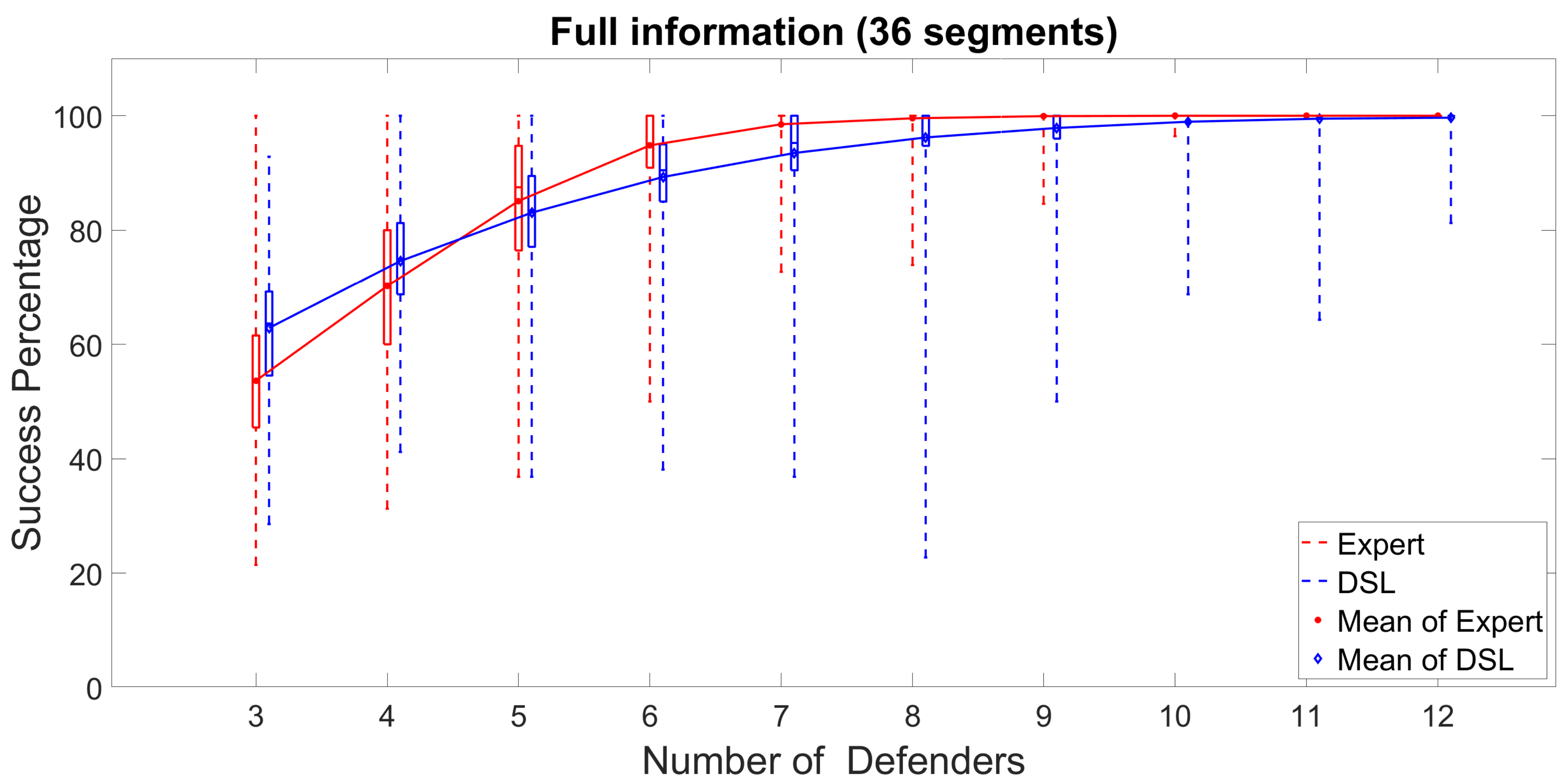}%
\label{fig:scalability_full}}
\caption{Scalability study of proposed DSL framework with different-sized team  of defenders (a)Partial observation scenario (b)Full observation scenario}
\label{fig:Scalability}
\end{figure*} 

\subsection{Comparison study}

An adaptive partitioning (AP) \cite{macharet2020adaptive} approach has been proposed for PDP for sequential capture of radially incoming intruders. The arrival distribution of the intruders is estimated to partition the perimeter into sectors; each defender assigned to these partitions independently operates to capture intruders sequentially. Although both AP and DSL are designed for different settings, we are comparing them to show their efficacy.
The naive and adaptive partitioning is used for the PDP, and the remaining approaches  are used for discrete PDP.  The dataset is generated with Poisson distribution with $ \lambda = 4$ and at most one intruder in each segment.  

 Table \ref{table:comparison_AP} shows the performance comparison for success percentage for adaptive partitioning, expert policy, and DSL framework. Note that, in \cite{velhal2022decentralized}, a comparison study showed that STMTA performs better than adaptive partitioning in the same setting.

\begin{table}[t!] 
\centering
\caption{ Average percentage and standard deviation of captured intruders for uniform arrival distribution of intruders, over 1000 runs for Naive, adaptive partitioning (AP), expert policy, and DSL framework.}
\label{table:comparison_AP}
 \setlength{\arrayrulewidth}{1pt}
\begin{adjustbox}{max width=0.95\linewidth} 
\renewcommand{\arraystretch}{1.2}
 \begin{tabular}{|l|r|}
\hline
 & \multicolumn{1}{c|}{Success percentage ($\mu \pm 3\sigma$))} \\ \hline
Naive   (full observation)   & $64.9 \pm 3.2$           \\ \hline
AP   (full observation)      & $64.9 \pm 3.2$        \\ \hline
Expert (full observation) &  $ 85.0425 \pm 4.8555 $     \\ \hline
DSL (full observation) &   $82.8825\pm  2.5617$    \\ \hline
Expert (partial observation) &  $85.3439 \pm 4.8887$      \\ \hline
DSL (partial observation) & $79.4924\pm 2.3139$    \\ \hline
\end{tabular}
\end{adjustbox}
 \end{table}

\subsection{Scalability of proposed DSL framework for d-PDP} 
 
In the proposed DSL framework once a defender is trained using MLC-SEFRON, the same network can be used for the different-sized team of defenders. Re-training is not required for changes in team size. Also the training of the defender is independent of the team size of defenders.  Hence, the  proposed DSL framework is said to be distributed and scalable. 

In this paper, the MLC-SEFRON is trained with a dataset generated using STMTA formulation proposed in \cite{velhal2022DREAM}, considering a team of five defenders. All different sized-team of defenders in the DSL framework will use the same network (which is trained with a team size of five).
The expert STMTA solution is computed (based on the actual number of defenders) using centralized computation.
Figure \ref{fig:Scalability} shows the performance of the proposed DSL framework with different team sizes.
One can observe that, the solution obtained from the STMTA approach and the DSL framework is comparable. Both the solutions show similar characteristics;  as the defenders' team size increases the success rate increases. 
One should note that the expert used for training (i.e. the solution of STMTA) is sub-optimal. The performance of DSL framework asymptotically  matches  the success rate of the expert.  With the increases in the number of defenders, the difference between the success rates reduces. Hence, the  proposed DSL framework is scalable.

One can observe that, the DSL framework performs better compared to the expert approach for team sizes of 3 and 4 in a partial observation scenario. 
It is an interesting result where the DSL solution outperforms the expert.  The DSL has learned to solve the assignment problem and not just the given expert solution (which is sub-optimal). Investigating the reasons for this observation is out of the scope of this paper.  If the sensing range of the defender changes then the number of neurons in the MLC-SEFRON network of the DSL framework also changes. The proposed DSL framework for d-PDP is not scalable in terms of the sensing range of the defender.

\section{Conclusion} \label{sec:conclusion}

In this paper, a novel Decentralized Spike-based Learning (DSL) framework for the discrete Perimeter Defense Problem (d-PDP) is proposed. The circular territory is divided into multiple segments and hence the PDP is termed as d-PDP. Initially, the d-PDP is formulated as a Spatio-Temporal Multi-Task Assignment problem (STMTA). Then this STMTA problem is converted into deterministic multi-label learning using Multi-Label Classifier using Synaptic Efficacy Function spiking neuRON (MLC-SEFRON) network. For decentralized training, each defender is trained with a separate MLC-SEFRON network. The spatiotemporal information of the defenders and intruders in the territory is converted as spikes without any extra pre-processing step. These spikes are given as input to the MLC-SEFRON network. The labels of segments assigned to the defender are obtained from the output of the MLC-SEFRON network The trained weights of one MLC-SEFRON network are directly used for obtaining labels of assigned segments from other MLC-SEFRON networks without any retraining. Using these labels the trajectories for defenders are then obtained with the help of the Consensus-Based Bundle Algorithm (CBBA). The performance of MLC-SEFRON is evaluated for full and partial observation scenarios of the defenders. The multi-label performance metrics obtained for MLC-SEFRON are observed to follow the input data which is obtained from the expert policy. The performance results show that the proposed DSL framework  performs $ 93.14\%$ and $ 97.5\%$ compared to the expert policy in partial and full observation scenarios, respectively. The DSL performance evaluated for different numbers of defenders indicates that it is efficiently scalable. Future work will explore improving the performance of the proposed MLC-SEFRON and its application for various multi-label learning as well as STMTA problems. The presented d-PDP dataset will be a good benchmark for testing the SNN learning algorithms due to its inherent spatio-temporal nature.

\bibliographystyle{IEEEtran}
\bibliography{PDP_SNN_bib.bib}

\end{document}